\def\year{2019}\relax
\documentclass[letterpaper]{article} %DO NOT CHANGE THIS
\usepackage{aaai19}  %Required
\usepackage{times}  %Required
\usepackage{helvet}  %Required
\usepackage{courier}  %Required
\usepackage{url}  %Required
\usepackage{graphicx}  %Required
\usepackage{amsmath}
\usepackage{subfigure}
\begin{document}
\title{Learning behavioral context recognition with multi-stream \\ temporal convolutional networks}
\author{Aaqib Saeed, Tanir Ozcelebi, *Stojan Trajanovski, Johan Lukkien \\
\small{a.saeed@tue.nl, t.ozcelebi@tue.nl, stojan.trajanovski@philips.com, j.j.lukkien@tue.nl} \\
\small{Department of Mathematics and Computer Science, Eindhoven University of Technology, Eindhoven, The Netherlands}\\
\small{*Philips Research, Eindhoven, The Netherlands}\\
}

\maketitle
\begin{abstract}
Smart devices of everyday use (such as smartphones and wearables) are increasingly integrated with sensors that provide immense amounts of information about a person's daily life such as behavior and context. The automatic and unobtrusive sensing of behavioral context can help develop solutions for assisted living, fitness tracking, sleep monitoring, and several other fields. Towards addressing this issue, we raise the question: can a machine learn to recognize a diverse set of contexts and activities in a real-life through joint learning from raw multi-modal signals (e.g. accelerometer, gyroscope and audio etc.)? In this paper, we propose a multi-stream temporal convolutional network to address the problem of multi-label behavioral context recognition. A four-stream network architecture handles learning from each modality with a contextualization module which incorporates extracted representations to infer a user's context. Our empirical evaluation suggests that a deep convolutional network trained end-to-end achieves an optimal recognition rate. Furthermore, the presented architecture can be extended to include similar sensors for performance improvements and handles missing modalities through multi-task learning without any manual feature engineering on highly imbalanced and sparsely labeled dataset.
\end{abstract}

\section{Introduction}
The problem of context recognition is centered on inferring person's environment, physical state, and activity performed at any particular time. Specifically, a understanding of the user's current context requires determining where and with whom the person is? and in what type of activity the person is involved in? The behavioral and activity analysis is an important and challenging task mainly because it is crucial for several applications, including smart homes~\cite{rashidi2009keeping}, assisted living~\cite{lin2015disorientation,rashidi2013survey}, fitness tracking~\cite{rabbi2015mybehavior}, sleep monitoring~\cite{lin2012bewell+}, user-adaptive services, social interaction~\cite{lee2013sociophone} and in industry. In particular, an accurate recognition of human context can greatly benefit healthcare and wellbeing through automatic monitoring and supervision of patients with chronic diseases~\cite{lara2013survey} such as hypertension, diabetes and dementia~\cite{ordonez2016deep}. Furthermore, the gathered knowledge and extracted activity patterns can enable novel treatment design, adjustment of medications, better behavioral intervention and patient observation strategies~\cite{lorincz2009mercury}. 

In practice, for a context detection system to be effective in a real-life requires an unobtrusive monitoring. It is important to not distress a person in order to capture their realistic behaviors in a natural environment. The penetration of smart sensing devices (e.g. smartphones and wearables) that are integrated with sophisticated sensors in our daily lives provides a great opportunity to learn and infer about various aspects of a person's daily life. However, there is considerable variability in the human behavior in real-world situations that can cause the system to fail, if it is developed using data collected in a constrained environment. For instance,~\citeauthor{miluzzo2008sensing} shows that the accuracy of activity classification differs based on the interaction with the phone e.g. when in hand or carried in the bag. The various sensors embedded in the smart devices convey information about different ambient facets each with a distinct prospect. The variability issues of different patterns in phone usage, environments, and device types can be very well addressed (to improve the recognition capability of the system) through learning disentangled representations from a large-scale data source and fusing rich sensory modalities rather than separately utilizing each of them. 
\begin{figure}
    \centering
    \includegraphics[width=8.0cm]{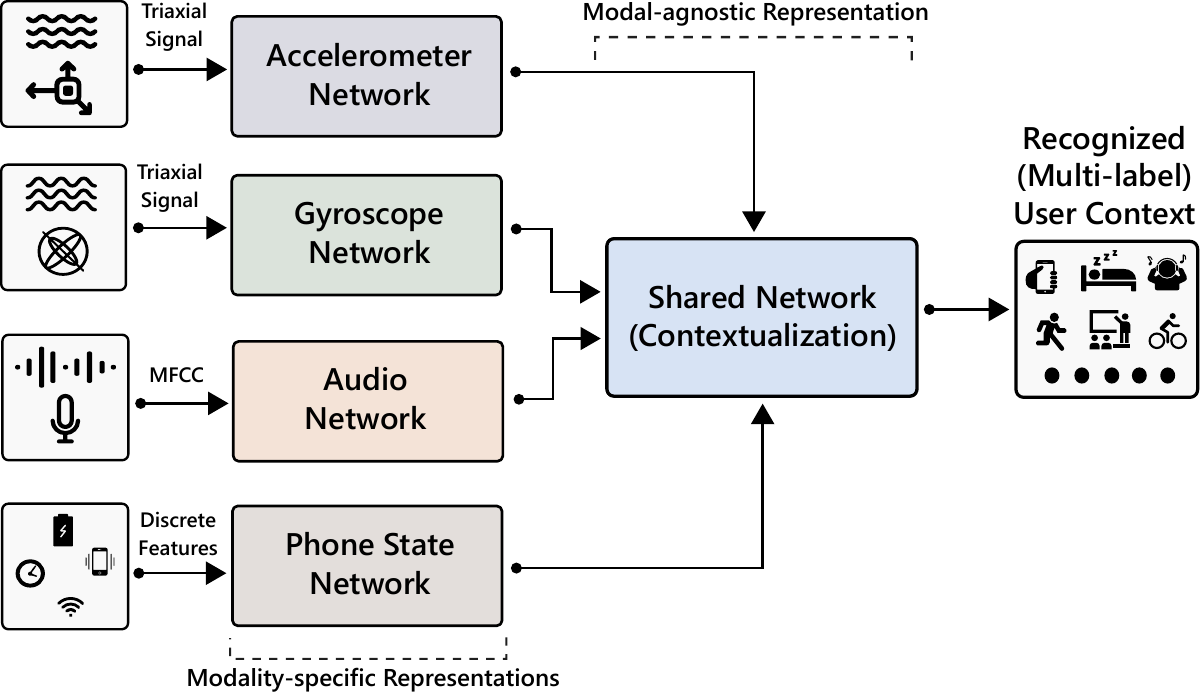}
    \caption{\textbf{Multi-modal representation learning from sensors:} \small{Schematic of the proposed multi-stream convolutional network.}}
    \label{fig:overview}
\end{figure}

\begin{figure*}[!htbp]
\centering
\subfigure[Audio (MFCC)]{\includegraphics[width=.33\textwidth]{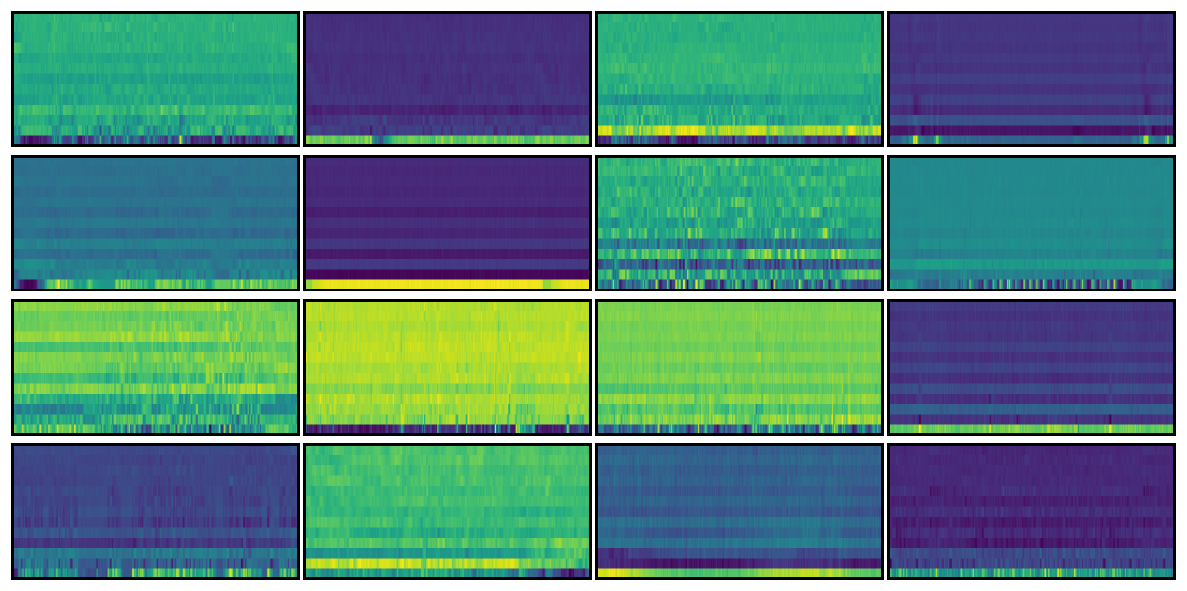}}
\subfigure[Accelerometer]{\includegraphics[width=.33\textwidth]{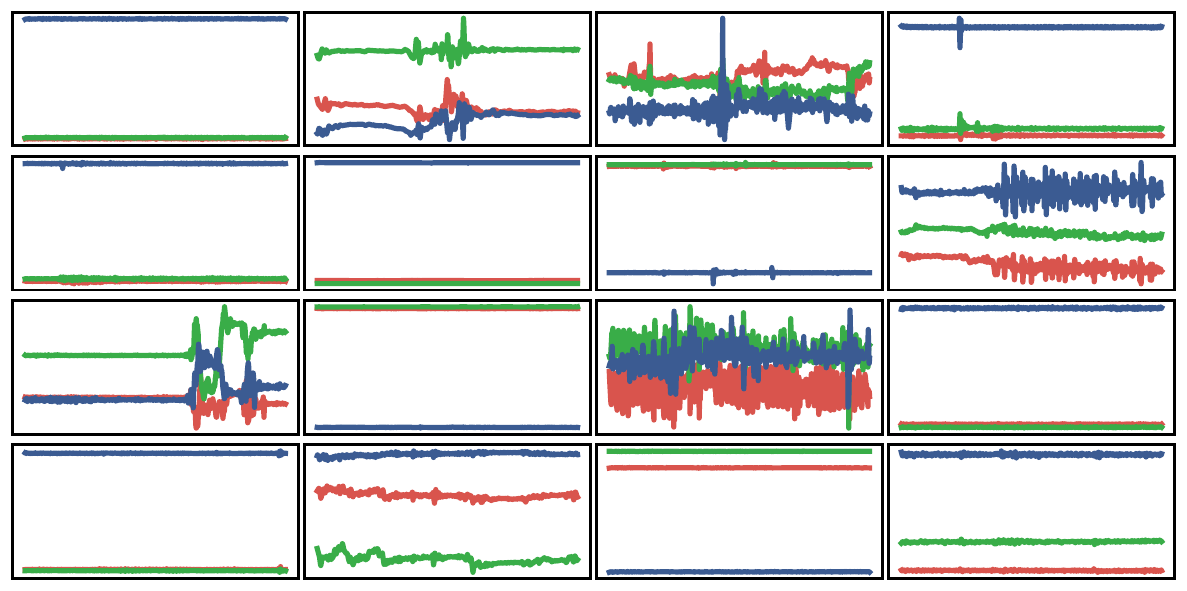}}
\subfigure[Gyroscope]{\includegraphics[width=.33\textwidth]{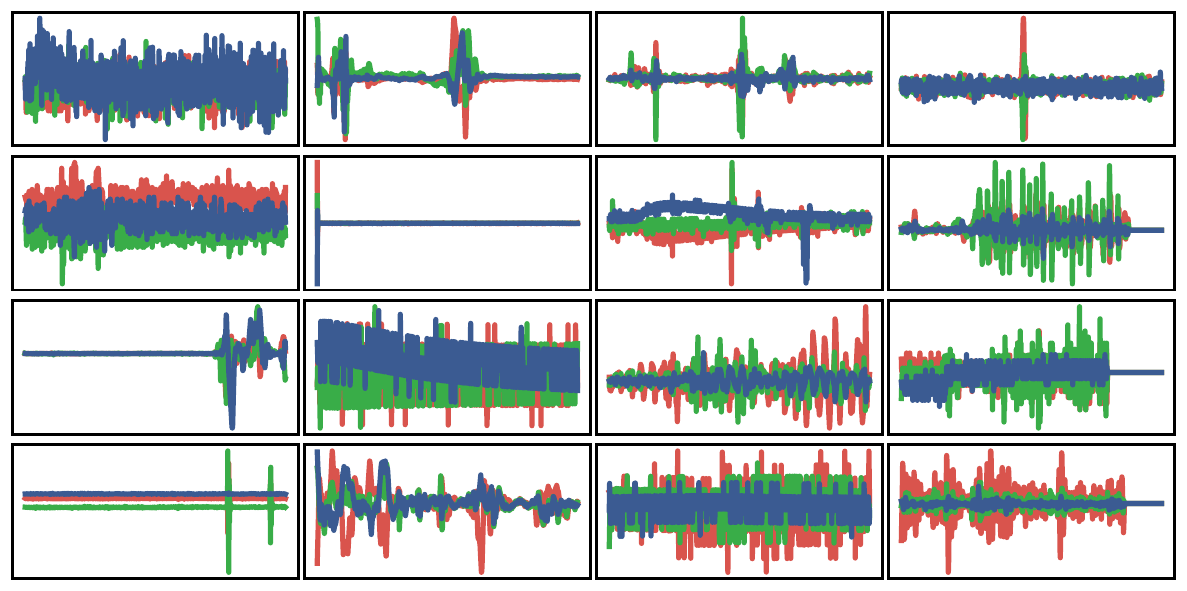}}
\caption{\textbf{Context recognition dataset:} \small{Samples from large-scale multi-modal sensory data collected in-the-wild conditions. The individual plots within each sub-figure correspond to the same set of activities/context.}}
\label{fig:samples}
\end{figure*}

In the past, several studies have shown great improvement in sensor processing for basic activity recognition~\cite{lara2013survey,hoseini2013survey}. The majority of the earlier methods use shallow learning classifiers (such as, Random Forest and Support Vector Machine) with hand-engineered features extracted from raw sensor readings e.g. heuristically selected statistical or frequency measures~\cite{figo2010preprocessing}. Likewise, many studies involve simulated controlled trials for data collection in lab environments that require users to wear extra sensors. Broadly, they also treat activity recognition as a multi-class classification problem, where a user's activity at a specific moment can be defined by one of the \textit{k} defined classes. On the contrary, people are not generally engaged in just one activity in their day-to-day living e.g. a person might surf the web while eating or talking to friends. These problems limit the applicability of these studies to detect very few rudimentary activities and make it harder for the system to generalize to real-life settings. Nevertheless, to be successful in everyday scenarios, the context recognition module should support a diverse set of activities, varying device usage, and a wide range of environments. Importantly, it must not only learn discriminative representations directly from raw signals without any ad-hoc feature engineering, but also seamlessly combine the discovered explanatory factors in the milieu of diverse sensory modalities~\cite{bengio2013representation}.

In recent years, the fields of speech recognition, drug discovery, image segmentation and machine translation have been tremendously revolutionized thanks to the availability of massive labeled datasets and end-to-end deep representation learning~\cite{bengio2013representation}. Similarly, the domain of human activity recognition has also started leveraging deep neural networks for automatic feature learning~\cite{ordonez2016deep,radu2018multimodal,yang2015deep} though commonly restricted to the detection of only elementary activities such as, walking, sitting, standing etc. There has not been the same progress in recognizing complex behavioral context in daily-life situations using devices of daily use. This can be partially attributed to the lack of a large labeled dataset, which is both expensive and time-consuming to accumulate in a real-world settings. We believe that large-scale sensory data can significantly advance context recognition. This issue is very recently addressed in~\cite{vaizman2017recognizing,vaizman2018context} which has open-sourced multi-modal data (see Figure~\ref{fig:samples}) of activities in-the-wild. The authors provide a baseline system for sensor fusion and a unified model for multi-label classification. They trained logistic regression and fully connected neural networks on hand-crafted features that are extracted based on extensive domain-knowledge. In this paper, we utilize this heterogeneous sensors data collected over a week from sixty users to learn rich representations in an end-to-end fashion for recognizing multi-label human behavioral context.

The  task of learning detailed human context is challenging, especially from imbalanced and multi-label data. Unconstrained device usage, a natural environment, different routines, and authentic behaviors are likely to result in a joint training dataset from several users with significant class skew~\cite{vaizman2018context} and missing labels. Another challenge with learning from multi-modal signals is developing an architecture that feasibly combines them as in diverse environments a certain sensor might perform better than others. For instance, if a person is watching a television with a phone lying on the table, the sound modality may dominate in the network as compared to an accelerometer. We address the former issue with instance weighting scheme same as ~\cite{vaizman2018context} and later through a unified architecture that can efficiently fuse representations in multiple ways.

We present a deep temporal convolutional neural network (CNN) that learns directly from various modalities through a multi-stream architecture (accelerometer, gyroscope, sound and phone state networks). Here, a separate network facilitates learning from each modality and a contextualization module incorporates all the available information to determine the user's context (see Figure~\ref{fig:overview}). In our experiments, we show that deep multi-modal representations learned through our network without any sophisticated pre-processing or manual feature extraction achieve state-of-the-art performance. 

The primary contribution of this paper is in showing how to leverage ample amount of raw sensory data to learn deep cross-modal representations for multi-label behavioral context. Although, the methods in the paper are standard, their application on a large-scale imbalanced and sparsely labeled smartphone data set is unique. The proposed network architecture achieves sensitivity and specificity score of $0.767$ and $0.733$, respectively averaged over $51$ labels and $5$-folds cross-validation. The rest of the paper describes our technique and experiments in detail. First, we review the related work on activity recognition. Then we present our multi-stream temporal convolutional network, architectural modifications for handling missing sensors, the proposed training procedure and implementation details. Next, the description of the dataset, evaluation protocol and experimental results are described, followed by the conclusions. 
\begin{figure*}[!htbp]
\centering
\includegraphics[width=14cm]{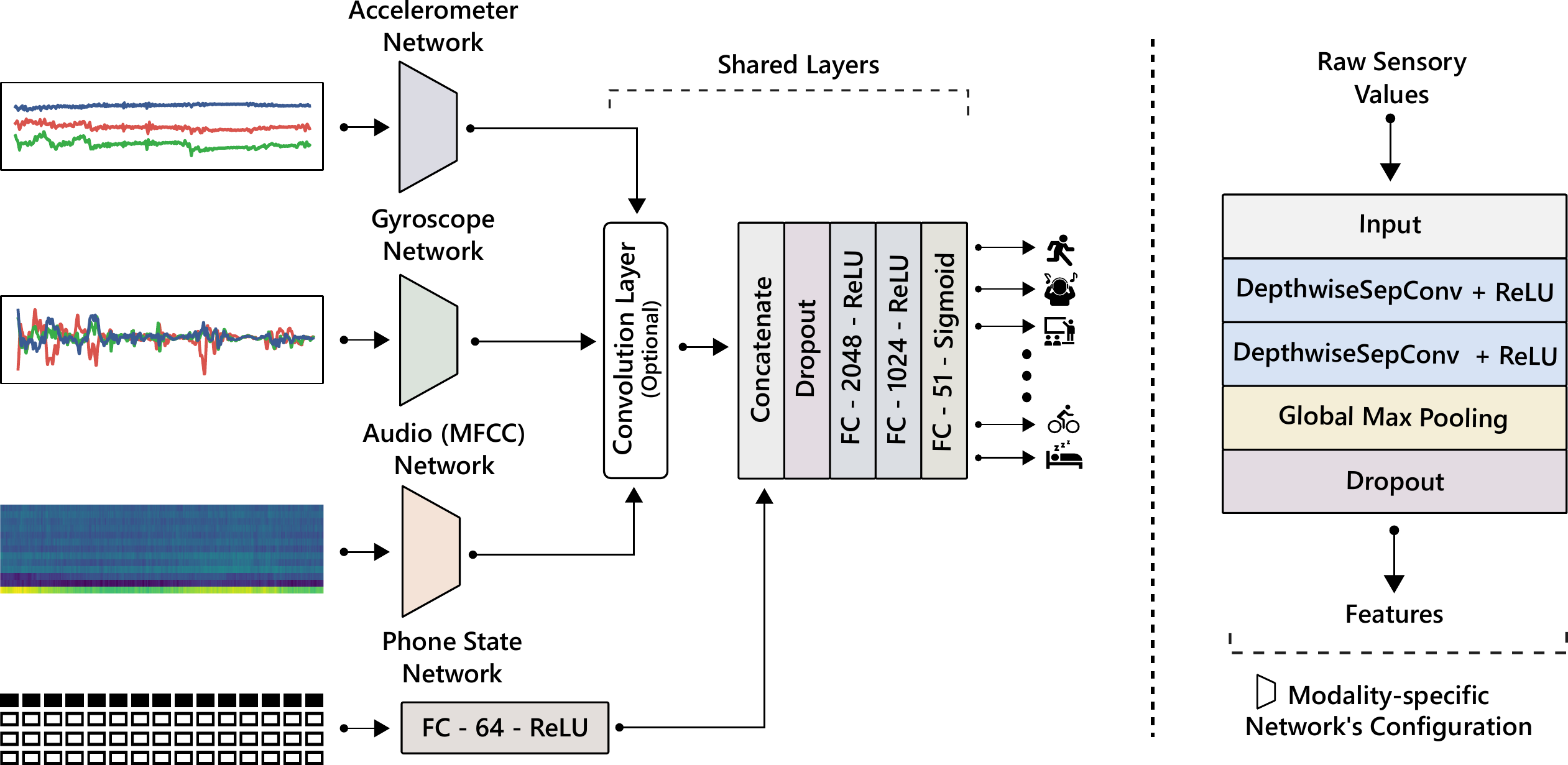}
\caption{\textbf{End-to-end multi-modal and multi-label context recognition:} \small{We propose a deep temporal convolutional architecture for multi-label behavioral context recognition. A separate network learns representations (features) from each modality using depthwise-separable convolutions and contextualizes this information through shared layers to infer the user context.}}
\label{fig:architecture}
\end{figure*}
\section{Related Work}
Human activity recognition has been extensively studied in simulated and controlled environments. It is concerned with classifying sensor measurements into existing activity categories. The earlier techniques are predominantly based on applying shallow learning algorithms on manually extracted features (e.g. statistical and spectral attributes)~\cite{figo2010preprocessing}. Despite there are unsupervised~\cite{bhattacharya2014using,plotz2011feature} and supervised~\cite{yang2015deep,ordonez2016deep,ronao2016human,zeng2014convolutional} deep learning methods applied for automatic feature extraction to detect activities, these approaches are fairly limited by the amount of labeled data (of many sensing modalities) from the real-world. Furthermore, they do not fully address the issue of multi-label context recognition. A user state is described by only one class or label, which is not true for activities humans perform in real-life. Moreover, only recently the exploration has begun into joint-learning and fusing multiple modalities for ubiquitous sensing through deep networks~\cite{radu2018multimodal,vaizman2018context}. The works cited here are by no means an exhaustive list, but provide a recent representative advancements made in utilizing deep neural networks for activity recognition. We recommend the interested readers to refer~\cite{rashidi2013survey,shoaib2015survey} for an extensive survey of former approaches. 

A systematic analysis of several deep neural architectures for activity recognition is provided by~\citeauthor{hammerla2016deep}. The suitability of various models is investigated that were trained only on raw accelerometer signals for different activity classification tasks. On diverse benchmark datasets, CNN and long-short-term memory networks are found to outperform hand-crafted features by a significant margin. Likewise,~\citeauthor{alsheikh2016deep} proposed an approach combining pre-training and fine-tuning of deep belief networks for sequential activity recognition. They extracted spectrograms from a triaxial accelerometer and found them to be helpful for capturing variations in the input. Similarly,~\citeauthor{jiang2015human} used $2$D activity images extracted from accelerometer signals as CNN input. The importance of unsupervised training of models in feature learning and optimization is highlighted in~\cite{bhattacharya2014using} using a combination of sparse-coding framework and semi-supervised learning. Likewise,~\citeauthor{yang2015deep}  developed a multi-channel CNN model to replace heuristic based hand-crafted features. Their analysis showed CNNs work well compared to traditional (shallow) learning algorithms on several datasets. Audio sensing is also employed in unconstrained acoustic environments through applying fully connected neural networks~\cite{lane2015deepear}. Recently,~\citeauthor{radu2018multimodal} used deep networks for multi-modal activity recognition and compared them with traditional learning algorithms on various recognition tasks. Likewise, numerous other studies also positively utilize deep learning for detection of basic activities~\cite{ordonez2016deep,ronao2016human,zeng2014convolutional}. 

We differentiate ourselves from the existing approaches through utilizing a deep multi-stream CNN (with depthwise separable convolutions) on a large and diverse context detection dataset. Specifically, we build on previous work by~\citeauthor{vaizman2018context} that only employed hand-engineered features for training linear and shallow neural networks. In contrast, our general-purpose approach allows us to train a deeper network that can not only automatically discover hidden latent factors, but also seamlessly combine them to achieve an end-to-end learning system without requiring domain expertise. Moreover, through taking advantage of multi-task learning~\cite{caruana1997multitask} we develop an architecture that can robustly handle missing sensors.

\section{Learning Multi-Modal Networks}
We design a deep convolutional neural network to address the problem of behavioral context recognition through learning representations from raw sensory inputs. To deal with cross-modality signals i.e. accelerometer (Acc), gyroscope (Gyro), audio (MFCC/Aud), and phone state (PS), we use a multi-stream architecture. The network comprises five main modules as demonstrated in Figure~\ref{fig:architecture}. This section describes each component, presents a strategy to modify the proposed architecture to handle missing sensors and provides the implementation details. 

\subsection{Modality Specific Networks}
We present a deep multi-modal convolutional architecture for learning context representations. We propose to use a series of depthwise-separable convolutions (DPS-Conv)~\cite{chollet2017xception} for processing different components (or channels) of raw signals. In general, CNNs are also found to be well suited for processing $1$D sequences due to their ability to learn translation invariant features, scale separation, and localization of filters across time and space~\cite{bai2018empirical}. DPS-Conv consists of two operations i.e. a depthwise convolution and a pointwise (or $1$ x $1$) convolution. Specifically, the first function (depthwise convolution) performs a convolution independently over each input channel and it is followed by the second operation of $1$ x $1$ convolution that projects the channels estimated by the earlier onto a distinct channel space to have the same number of output filters~\cite{kaiser2017depthwise}. The intuition of this formulation falls in line with the classical procedures utilized by domain experts to extract several features from each signal component independently (e.g. $x$, $y$ and $z$ constituents of an accelerometer) but pointwise convolution goes one step further and tries to learn unified factors that may capture relationships among independent elements. Moreover, separable convolutions make efficient use of parameters as opposed to their classical counterpart and this property has made them a very promising candidate for contemporary architectures that run on smart devices with limited computing and energy capabilities~\cite{sandler2018mobilenetv2,zhang2017shufflenet}. Formally, in case of $1$D input sequence $\mathbf{x}$ of length $L$ with $M$ channels, the aforementioned operation can be formulated as follows~\cite{kaiser2017depthwise}:  
\begin{equation*}
    \text{DepthwiseConv}(\mathbf{x}, \mathbf{w})_{i} = \sum_{l}^{L} (\mathbf{x}[i:i + k - 1] \odot \mathbf{w})_l
\end{equation*}
\begin{equation*}
    \text{PointwiseConv}(\mathbf{x}, \mathbf{w})_{i} = \sum_{m}^{M} (\mathbf{x}[i:i + k - 1] \cdot \mathbf{w})_m
\end{equation*}
\begin{multline*}
    \text{DepthwiseSeparableConv}(\mathbf{x}, \mathbf{w_d}, \mathbf{w_p})_{i} =  \\ \text{PointwiseConv}_{i}( \text{DepthwiseConv}_{i} (\mathbf{x}[i:i + k - 1], \mathbf{w_d}), \mathbf{w_p}\\
\end{multline*}
\noindent where $\odot$ is elements-wise product, $\mathbf{x}[i:j]$ represents a segment of the complete sequence with adjacent columns from $i$ to $j$, and $\mathbf{w}$ represents filter with receptive field size of $k$. 

The proposed network takes four different signals as input, each with its independent disjoint pathway in the earlier layers of the network. Towards the end, they are merged into shared layers that are common across all modalities  that are described in the next subsection. This network configuration has the benefit of not just extracting modality-specific (and channel-specific) features but it can also feasibly extract mutual representations through shared layers. Each of the presented Acc and Gyro networks consist of $2$ temporal convolution layers which act as feature extractors over raw signals of dimensions $800$ x $3$. The convolution layers have kernel sizes of $64$ and $32$ with a stride of $2$ and each layer has $32$ and $64$ filters, respectively. We use rectified linear activation in all the layers and apply depth-wise L$2$-regularization with a rate of $0.0001$. The audio network takes mel frequency cepstral coefficients (see Section Dataset and Modalities) of size 420 x 13 as input and it has a similar architecture except the kernel size, which is set to $8$ and $6$ in the first and second layers, respectively. Likewise, the discrete attributes indicating PS are fed into a single layer fully-connected (FC) network with $64$ units and L$1$-penalty is used on the weights with a rate of $0.0001$. Furthermore, we explore different mechanisms to get a fixed dimension vector from each modality that can be fed into a shared network. Specifically, we use: a) global max pooling (GMP), b) global average pooling (GAP), c) a FC layer,  and d) exactly pass the representations without any transformation to the shared network. 

\subsection{Shared Network (Contextualization)}
Given the concepts extracted from each modality, the shared network generates a modal-agnostic representation. To achieve this, we fuse the output of earlier networks either through concatenation or apply standard convolution (only for Acc, Gyro and Aud). We then feed the output into $2$ FC layers having $2048$, $1024$ hidden units, respectively. Same as earlier, we use rectified linear non-linearity and L$1$-regularization with a weight decay coefficient of $0.0001$. The final output layer contains $51$ units (one for each label) with sigmoid activation. Figure~\ref{fig:architecture} visualizes the sharing of the network layers, where, earlier layers are modality specific but downstream layers become more general. 

\subsection{Missing Sensors}\label{sec:ms}
\begin{figure}[htbp]
    \centering
    \includegraphics[width=8cm]{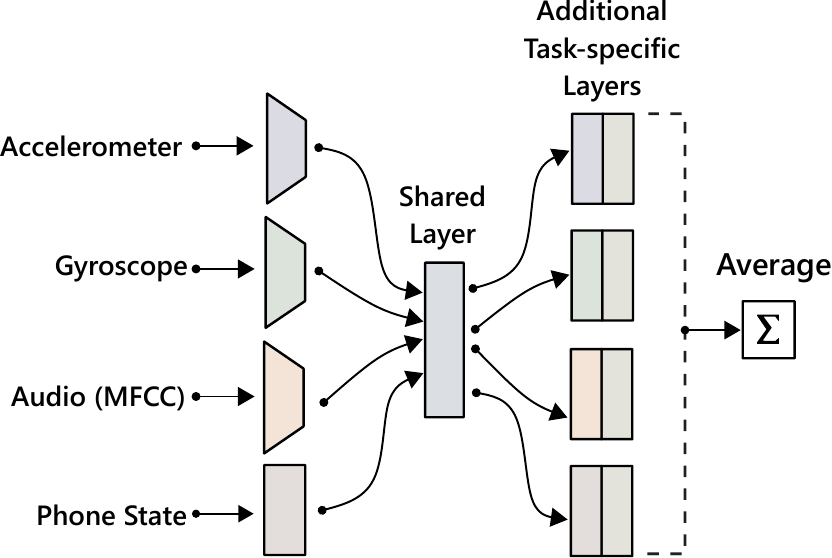}
    \caption{\textbf{Handling Missing Sensors with a Multi-task Network:} \small{A variant of the earlier defined architecture with additional task (modality-specific) layers and a separate loss function for each modality. It is able to recognize user context even if only one sensor is producing data and the others are unavailable.}}
    \label{fig:mmcnn_ms}
\end{figure}

In a real-life setting, a context recognition system may encounter missing modalities which can limit its inference capability. To make the model robust against such a situation, we develop a multi-task network~\cite{caruana1997multitask}, where learning from each sensor is posed as a task. The initial configuration of the model is the same as before but an additional layer (of $128$ units for Acc, Gyro, MFCC/Aud and $64$ units for PS) with a separate loss function is added after only a single shared layer of $1024$ hidden units. Figure~\ref{fig:mmcnn_ms} provides a high-level overview of the architecture. We employ joint-training (with a learning rate of $0.0003$) on all the modalities through aggregating cost functions of each model in order to get a total loss. This architectural configuration allows not only to learn independent and shared factors but enables inference even when any of the sensors is missing. It does so through averaging (which can be weighted) over probabilities produced by the individual networks.   
\subsection{Implementation and Training Details}
The networks are implemented in Tensorflow~\cite{abadi2016tensorflow} and the models are learned from scratch; initializing the weights with Xavier technique~\cite{glorot2010understanding}. Dropout~\cite{srivastava2014dropout} is applied on the hidden layers with a probability of  $0.2$. We use the Adam optimizer with a learning rate of $0.0001$ (unless mentioned otherwise) and use a batch size of $100$. We optimize the model weights for a fixed number of iterations (i.e. $15000$) with mini-batch stochastic gradient descent and backpropagation using instance-weighted cross-entropy objective function:
% and backpropagation~\cite{lecun1998gradient} 
\begin{equation*}
    %\mathcal{J}_c = \min \limits_{\theta} E_{(X,Y)} \big[ \mathbf{1}^{T} \cdot \Psi \odot \mathcal{L}_{ce}(f_c(\mathbf{x}; \theta),\mathbf{y}) \big] %\circ
    \mathcal{J}_C = \dfrac{1}{NC} \sum_{i=1}^{N} \sum_{c=1}^{C} \Psi_{i, c} \cdot \mathcal{L}_{CE}(\hat{y}_{i,c},y_{i,c})
\end{equation*}
\begin{equation*}
    \mathcal{L}_{ce}(\hat{y},y) = - [(y\log(\hat{y}) + (1 - y)\log(1 - \hat{y}))]
\end{equation*}
\noindent where $\mathcal{L}_{ce}$ is the binary cross-entropy loss, and $\Psi$ is an instance-weighting matrix of size $N$ x $C$ (i.e. number of training examples and total labels, respectively). The instance weights in $\Psi$ are assigned by inverse class frequency. Likewise, the entries for the missing labels are set to zero, to impose no contribution in the overall cost from such examples. 

\section{Experimental Results}
We conduct several experiments to analyze the capability of the proposed method. First, we provide a brief description of the utilized dataset and signals. Second, we describe the evaluation approach and metrics used to determine the model's performance on a multi-label and imbalanced dataset. Finally, we discuss our empirical observations, effect of different modalities' representation, comparison of various procedures to learn shared factors and visualization of the internal representation.

\subsection{Dataset and Modalities}
We choose to learn discriminative representations directly from raw Acc, Gyro, Aud/MFCC and PS attributes from a smartphone because of their wide adoptability and ubiquity. For this purpose, we chose to leverage \textit{ExtraSensory Dataset}~\cite{vaizman2017recognizing} since it is collected in a natural environment from users' personal devices. The experimental setup was not scripted but data collection was performed when participants were busy with their daily routines to capture varied activities and context combinations, in-the-wild conditions. This data source contains over $300,000$ multi-labeled instances (with classes such as  `outside', `at a restaurant', `with friends' from a total of $51$ labels) from sixty users. The complete data collection protocol is described in~\cite{vaizman2017recognizing}. Here, we provide a high-level overview of the signals that we used in this study. The samples are collected for $20$ seconds duration every minute from tri-axis Acc and Gyro at a sampling frequency of $40$Hz, mel frequency cepstral coefficients (MFCCs) for $46$msec frame are extracted from Aud recorded at $22,050$Hz. Likewise, several phone state binary features are also collected such as those specifying, time of day, battery level, ringer mode and Wi-Fi connection etc. A few randomly selected samples of these signals are illustrated in Figure~\ref{fig:samples}. 

We seek to process raw sensory values without manual feature engineering. Thus, the only pre-processing we applied is to transform variable length inputs to an identical temporal length. For this purpose, the MFCCs of environmental audio are repeated (along time dimension) to get equal size input, this is reasonable for ambient soundscapes as we are not particularly interested in inferring a specific sound event. Similarly, the Acc and Gyro samples of varying sizes are zero-padded and instances, where MFCC length is shorter than twenty are discarded. Furthermore, we treat Acc, Gyro and Aud as $m$-channels inputs ($3$, $3$, and $13$ channels, respectively) as it allows us to efficiently learn independent factors from every sensor axis, thus maximally utilizing the large-scale dataset. 
\subsection{Evaluation and Metrics}
Our models are evaluated with five-folds cross-validation with the same divisions of sixty users as of~\cite{vaizman2018context}, where training and test folds contain $48$ and $12$ users, respectively. For hyper-parameter optimization, we use nested cross-validation~\cite{cawley2010over} by randomly dividing the training fold data into training and validation sets with ratio of $80$-$20$. After hyper-parameters selection, we train our models on the complete dataset of training folds (individually, each time from scratch) and calculate metrics on the testing folds. Furthermore, it is mentioned earlier that the considered dataset is highly imbalanced with sparse labels. In this case, simply calculating naive accuracy will be misleading due to not taking underrepresented classes into account. Similarly, precision and f$1$-score are also very likely to be affected by the class-skew due to involvement of true positives in the denominator. Hence, we adopt a metric named balanced accuracy (BA)~\cite{brodersen2010balanced} as used in~\cite{vaizman2018context}, which incorporates both recall (or true positive rate) and true negative rate: $\text{BA} = \frac{\text{Sensitivity} + \text{Specificity}}{2}$. BA can be interpreted as average accuracy achieved on either class (positive or negative regarding binary classification). It stays identical to traditional accuracy, if a model performs equally well on each class but drops to a random chance (i.e. $50$\%) if a classifier performs poorly on a class with few instances~\cite{brodersen2010balanced}. We calculate BA for each label independently and average them afterwards to get a trustworthy score of the model's overall performance.

\subsection{Results and Analysis} 
\subsubsection{Analysis of Fusing Multi-Modal Representations:} 
\begin{figure}[htb]
\centering
\includegraphics[width=8cm]{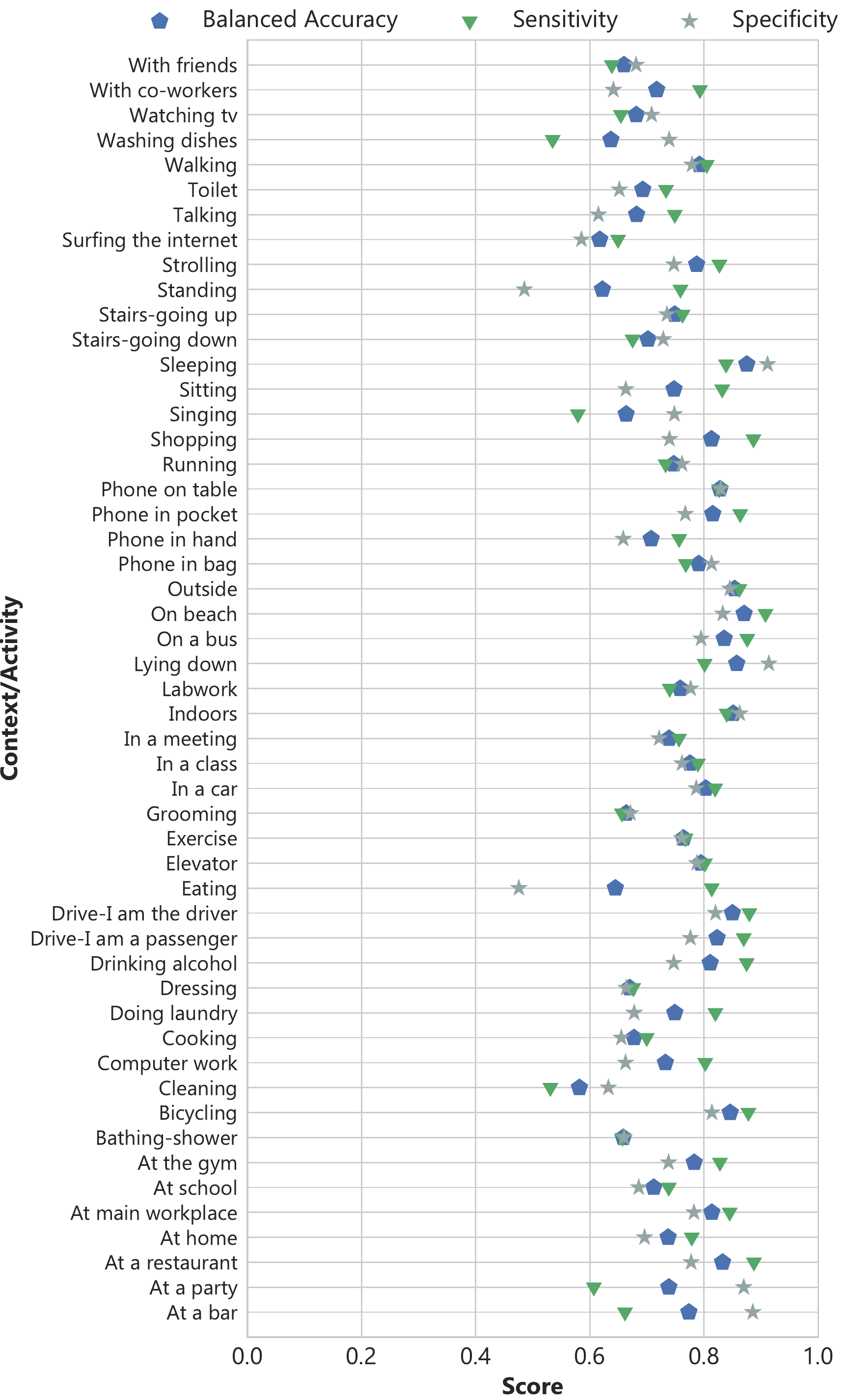}
\caption{\textbf{Performance metrics per label of the best performing model (with GMP):} \small{The scores are averaged over $5$-folds cross-validation.}}
\label{fig:smr}
\end{figure}

We quantify the effect of different procedures for getting a fixed dimension feature vector from each modality-specific network and examine their fusion through different configurations of the shared network. It is important to note that, we keep an entire network's configuration same but only the layers under consideration are changed. Table~\ref{tab:mf} provides the averaged (metrics) scores over $51$ contextual labels and $5$-folds as a result of applying global (max and average) pooling, using FC layer or simply feeding the extracted representations to the shared network for further processing. For the latter, we explore learning mutual representation from Acc, Gyr, and Aud/MFCC through an additional standard convolution layer and compare its performance with directly using flattened representations. Our experiments suggest that global max pooling (GMP) over each modality's features outperforms other utilized techniques; achieving BA of $0.750$ with a sensitivity rate of $0.767$. We believe the reason is that, GMP is capable of picking-up high-level shift-invariant features, which are most discriminative among others. Figure~\ref{fig:smr} presents per label metrics for this network on all the $51$ labels in the dataset. Specifically, we notice majority of the labels have BA score in range of $70\%$-$80\%$. 

\begin{table}[!htbp]
\centering
\caption{\textbf{Multi-modal context recognition:} The metrics are reported for $5$-folds cross-validation averaged over $51$ class labels. BA stands for balanced accuracy.}
\label{tab:mf}
\small{
\begin{tabular}{l|cccc}
            & \multicolumn{1}{c}{\textbf{BA}}         & \multicolumn{1}{c}{\textbf{Sensitivity}}         & \multicolumn{1}{c}{\textbf{Specificity}}  \\ \hline
GMP         & \multicolumn{1}{c}{0.750 \tiny{($\pm$ 0.012)}} & \multicolumn{1}{c}{0.767 \tiny{($\pm$ 0.015)}} & \multicolumn{1}{c}{0.733 \tiny{($\pm$ 0.016)}}  \\
GAP         & \multicolumn{1}{c}{0.748 \tiny{($\pm$ 0.009)}} & \multicolumn{1}{c}{0.753 \tiny{($\pm$ 0.012)}} & \multicolumn{1}{c}{0.742 \tiny{($\pm$ 0.015)}} \\
FC          & 0.744 \tiny{($\pm$ 0.009)}                     & 0.735 \tiny{($\pm$ 0.014)}                    & 0.753 \tiny{($\pm$ 0.008)}  \\
Flattened   & 0.742 \tiny{($\pm$ 0.014)}                     & 0.734 \tiny{($\pm$ 0.029)}                    & 0.749 \tiny{($\pm$ 0.007)}   \\
Conv & 0.738 \tiny{($\pm$ 0.011)}                     & 0.725 \tiny{($\pm$ 0.022)}                    & 0.752 \tiny{($\pm$ 0.022)}            
\end{tabular}
}
\end{table}

\subsubsection{Comparison of Convolution Variants:} We evaluate the complete multi-stream model through replacing only DPS-Conv layers with standard convolution (Std-Conv) in modality-specific networks. We did not observe major performance differences between the two models as shown in Table~\ref{tab:ccl}. Nevertheless, a model with DPS-Conv should be preferred because of having lower computational cost than Std-Conv~\cite{sandler2018mobilenetv2}.
\begin{table}[!htbp]
\centering
\caption{\textbf{Performance evaluation with different convolution layers.}}
\label{tab:ccl}
\small{
\begin{tabular}{l|ccc}
 & \textbf{BA} & \textbf{Sensitivity} & \textbf{Specificity} \\ \hline
Std-Conv & 0.751 \tiny{($\pm$ 0.011)} & 0.750 \tiny{($\pm$ 0.017)} & 0.751 \tiny{($\pm$ 0.007)} \\
\begin{tabular}[l]{@{}l@{}}DPS-Conv\end{tabular} & 0.750 \tiny{($\pm$ 0.012)} & 0.767 \tiny{($\pm$ 0.015)} & 0.733 \tiny{($\pm$ 0.016)}
\end{tabular}
}
\end{table}
\subsubsection{Quantifying Modality Influence:} To examine the effect of different combinations of sensors (or features learned from them) on the recognition capability of the model, we experimented with training several networks with modified architectures. Specifically, in this case the model only consisted of layers that are relevant to the signals under consideration e.g. for evaluating models with only Acc, Aud, and PS, we removed the Gyro network entirely and then trained it end-to-end from scratch. Table~\ref{tab:mi} shows the evaluation results that highlights the importance of joint-learning and fusion of multiple modalities to improve detection rate.    
\begin{table}[htbp]
\centering
\caption{\textbf{Effect of different modalities on recognition performance.}}
\label{tab:mi}
\small{
\begin{tabular}{l|cccc}
 & \textbf{BA} & \textbf{Sensitivity} & \textbf{Specificity} \\ \hline
Acc & 0.633 \tiny{($\pm$ 0.011)} & 0.668 \tiny{($\pm$ 0.027)} & 0.599 \tiny{($\pm$ 0.017)} \\
Gyro & 0.639 \tiny{($\pm$ 0.011)} & 0.638 \tiny{($\pm$ 0.017)} & 0.640 \tiny{($\pm$ 0.020)} \\
Aud & 0.669 \tiny{($\pm$ 0.024)} & 0.731 \tiny{($\pm$ 0.028)} & 0.608 \tiny{($\pm$ 0.025)} \\
PS & 0.712  \tiny{($\pm$ 0.005)} & 0.723 \tiny{($\pm$ 0.011)} & 0.700 \tiny{($\pm$ 0.013)} \\
Acc, Gyro, PS & 0.733 \tiny{($\pm$ 0.010)} & 0.744 \tiny{($\pm$ 0.021)} & 0.722 \tiny{($\pm$ 0.014)} \\
Acc, Gyro, Aud & 0.708 \tiny{($\pm$ 0.010)} & 0.722 \tiny{($\pm$ 0.027)} & 0.693 \tiny{($\pm$ 0.012)} \\
Acc, Aud, PS & 0.745 \tiny{($\pm$ 0.013)} & 0.757 \tiny{($\pm$ 0.025)} & 0.733 \tiny{($\pm$ 0.015)} \\
Gyro, Aud, PS & 0.748 \tiny{($\pm$ 0.012)} & 0.768 \tiny{($\pm$ 0.014)} & 0.728 \tiny{($\pm$ 0.014)} \\ \hline
All & 0.750 \tiny{($\pm$ 0.012)} & 0.767 \tiny{($\pm$ 0.015)} & 0.733 \tiny{($\pm$ 0.016)}
\end{tabular}
}
\end{table}

\subsubsection{Fusion and Effect of Missing Sensors:} 
We now evaluate the modified architecture's predictive performance (presented in Section Missing Sensors), confronting various combinations of missing signals. Table~\ref{tab:mms} provides experimental results showing that the proposed multi-task network can handle lost modalities, achieving similar BA score as when separate models for each modality are developed (see Table~\ref{tab:mi}). However, this flexibility comes at the price of slightly lower BA but makes a model capable of operation in the face of unavailable sensors.
\begin{table}[htbp]
\centering
\caption{\textbf{Assessment of multi-task network for handling missing modalities.} Each row provide averaged metrics score as earlier but only mentioned modalities that are used for determining user's context.}
\label{tab:mms}
\small{
\begin{tabular}{l|cccc}
\multicolumn{1}{c|}{} & \textbf{BA}  & \textbf{SN}  & \textbf{SP}   \\ \hline
Acc                   & 0.634  \tiny{($\pm$ 0.008)} & 0.652  \tiny{($\pm$ 0.027)} & 0.616  \tiny{($\pm$ 0.013)} \\
Gyro                  & 0.619  \tiny{($\pm$ 0.016)} & 0.632  \tiny{($\pm$ 0.040)} & 0.606  \tiny{($\pm$ 0.023)} \\
Aud                   & 0.656  \tiny{($\pm$ 0.026)} & 0.670  \tiny{($\pm$ 0.046)} & 0.641  \tiny{($\pm$ 0.015)} \\
PS                    & 0.688  \tiny{($\pm$ 0.009)} & 0.709  \tiny{($\pm$ 0.015)} & 0.667  \tiny{($\pm$ 0.012)} \\
Acc, Gyro             & 0.646  \tiny{($\pm$ 0.009)} & 0.670  \tiny{($\pm$ 0.028)} & 0.622  \tiny{($\pm$ 0.018)} \\
Acc, Aud              & 0.687  \tiny{($\pm$ 0.015)} & 0.695  \tiny{($\pm$ 0.035)} & 0.679  \tiny{($\pm$ 0.008)} \\
Acc, PS               & 0.708  \tiny{($\pm$ 0.007)} & 0.713  \tiny{($\pm$ 0.015)} & 0.702  \tiny{($\pm$ 0.012)} \\
Gyro, Aud             & 0.687  \tiny{($\pm$ 0.020)} & 0.699  \tiny{($\pm$ 0.045)} & 0.676  \tiny{($\pm$ 0.015)} \\
Gyro, PS              & 0.708  \tiny{($\pm$ 0.007)} & 0.719  \tiny{($\pm$ 0.023)} & 0.696  \tiny{($\pm$ 0.019)} \\
Aud, PS               & 0.708  \tiny{($\pm$ 0.013)} & 0.717  \tiny{($\pm$ 0.027)} & 0.698  \tiny{($\pm$ 0.010)} \\
Acc, Gyro, Aud        & 0.690  \tiny{($\pm$ 0.012)} & 0.703  \tiny{($\pm$ 0.031)} & 0.677  \tiny{($\pm$ 0.011)} \\
Acc, Gyro, PS         & 0.705  \tiny{($\pm$ 0.007)} & 0.714  \tiny{($\pm$ 0.023)} & 0.696  \tiny{($\pm$ 0.019)} \\
Acc, Aud, PS          & 0.721  \tiny{($\pm$ 0.007)} & 0.729  \tiny{($\pm$ 0.019)} & 0.712  \tiny{($\pm$ 0.011)} \\
Gyro, Aud, PS         & 0.721  \tiny{($\pm$ 0.011)} & 0.730  \tiny{($\pm$ 0.030)} & 0.711  \tiny{($\pm$ 0.017)} \\ \hline
All                   & 0.720  \tiny{($\pm$ 0.008)} & 0.728  \tiny{($\pm$ 0.025)} & 0.712  \tiny{($\pm$ 0.015)} 
\end{tabular}
}
\end{table}

\begin{figure}[tbp]
    \centering
    \includegraphics[width=7.4cm]{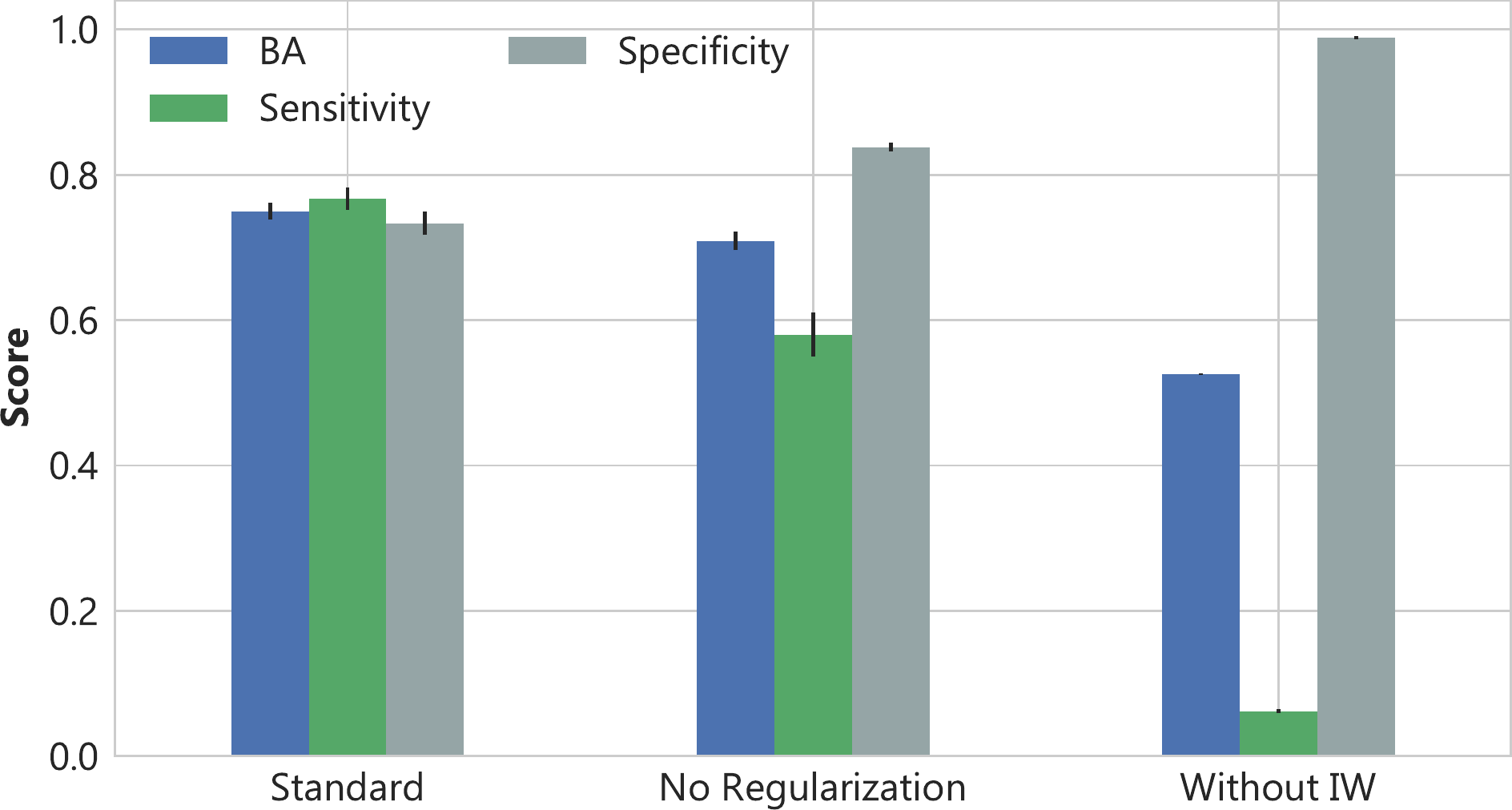}
    \caption{\textbf{Assessment of instance-weighting and regularization: } \small{We determine the impact of cost sensitive loss function and regularization (i.e. weight decay and dropout) on the network's predictive power. The results labeled under standard are with both IW and regularization.}}
    \label{fig:nriw}
\end{figure}

\begin{figure*}[htbp]
\centering
\subfigure[Environment]{\includegraphics[width=.24\textwidth]{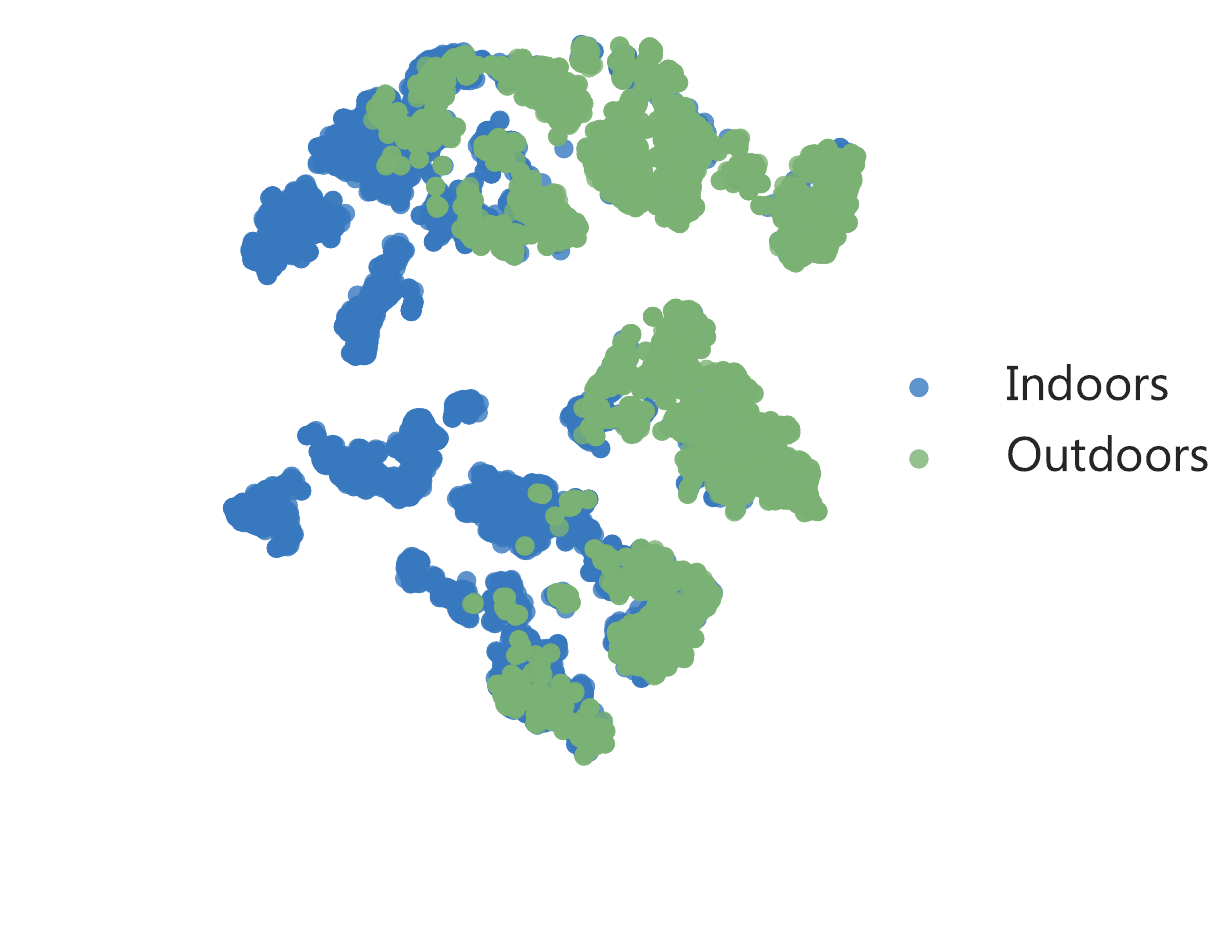}}
\subfigure[Body State]{\includegraphics[width=.24\textwidth]{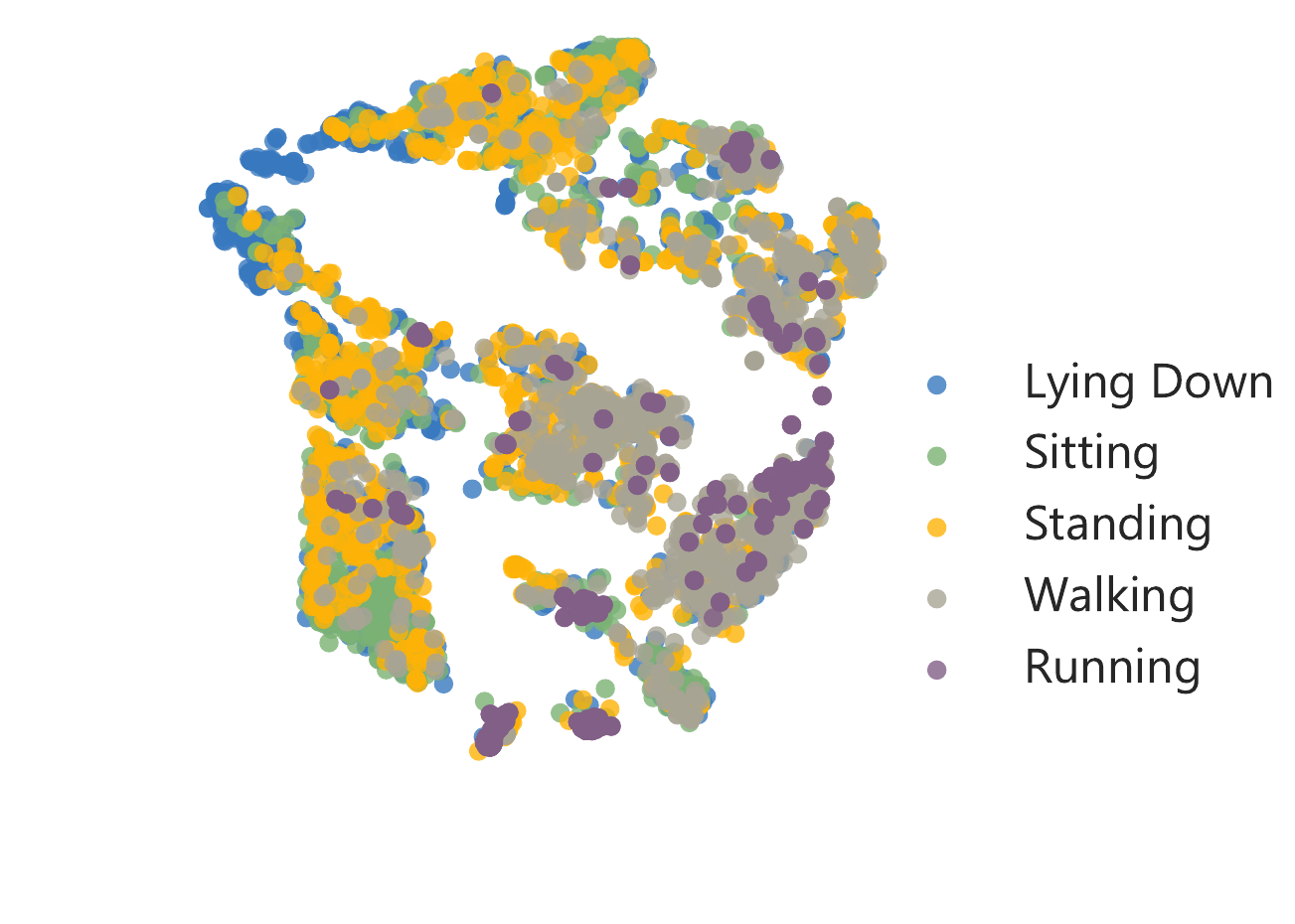}}
\subfigure[Transportation Mode]{\includegraphics[width=.24\textwidth]{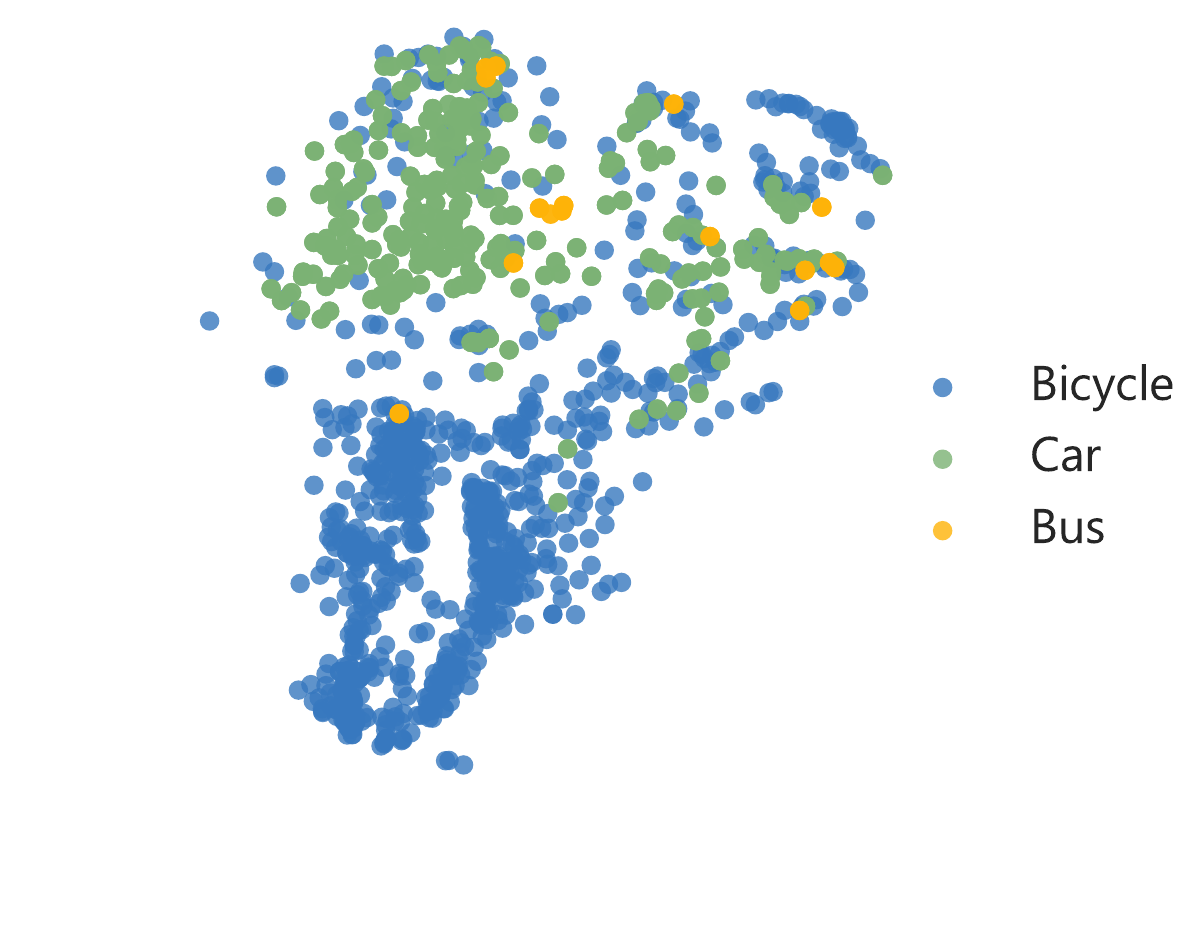}}
\subfigure[Phone Position]{\includegraphics[width=.24\textwidth]{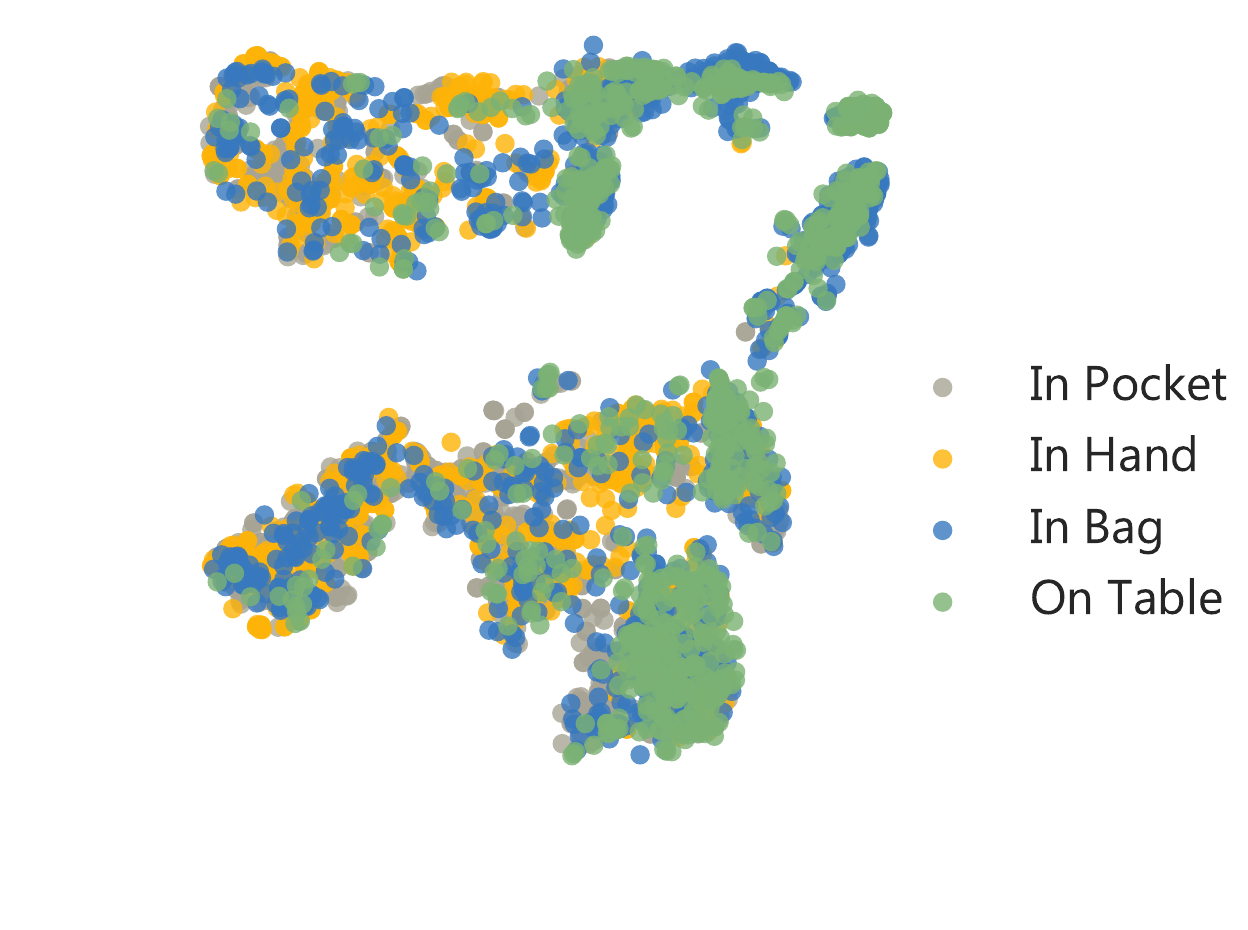}}
\caption{\textbf{t-SNE embeddings:} \small{We visualize the mutual features learned through fusion of multiple modalities (from the last layer) in the shared network. Four sets of mutually-exclusive labels are identified from multi-labeled data to use during final visualization of semantically related clusters extracted through t-SNE.}}
\label{fig:tsne}
\end{figure*}

\subsubsection{Reliance on Instance Weighting and Regularization:} Our results thus far have been obtained through training a model with cross-entropy loss. This incorporated instance-weights to handle class-imbalance. To test network's dependence on the cost sensitive loss function ($\mathcal{J}_c$), we examined a model's performance that is trained without it. As expected, the overall BA score drastically drops to a random chance (see Figure~\ref{fig:nriw}) with worse performance on positive samples in comparison with the negative ones. Likewise, we also trained a model without any sort of regularization i.e. removing dropout, L$1$ and L$2$ penalties from the network. The average recall rate on the held-out testing folds dropped to $0.58$ which can be an indication of overfitting the training set. Hence, incorporating both instance-weighting (IW) and regularization improved performance significantly in learning from this imbalanced dataset. However, further work will be necessary to investigate other techniques for managing (sparse) rare labels such as oversampling and data augmentation in case of multi-labeled instances. 

\subsubsection{Visualization:} 
\begin{figure}[htbp]
\centering
\subfigure[]{\includegraphics[width=.35\textwidth]{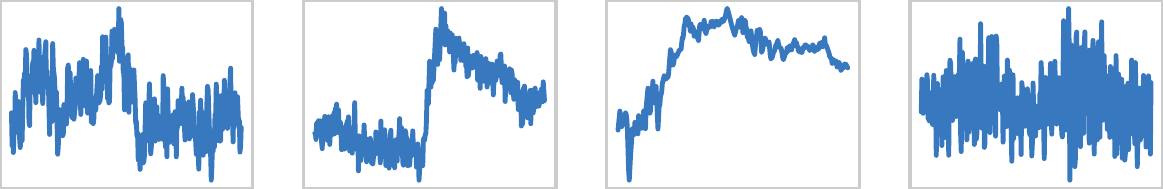}}
\subfigure[]{\includegraphics[width=.35\textwidth]{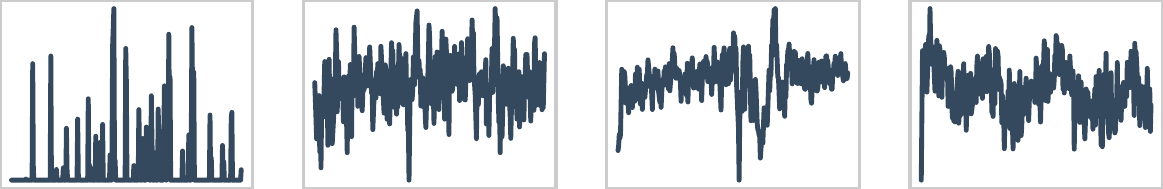}}
\subfigure[]{\includegraphics[width=.35\textwidth]{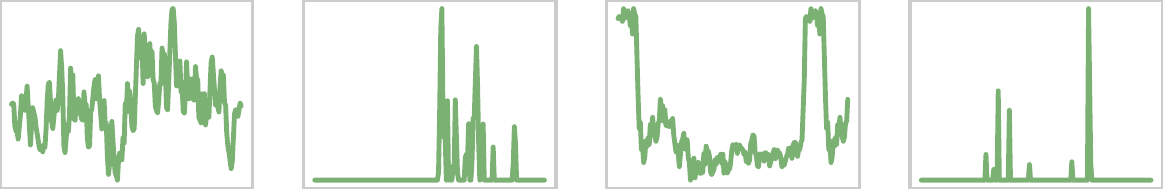}}
\caption{\textbf{Feature Maps from Modality-Specific Networks:} \small{Illustration of randomly selected (learned) features from first layer of convolutional networks. (a), (b) and (c) represent outputs from Acc, Gyro and Aud models, respectively.}}
\label{fig:fm}
\end{figure}

In order to illustrate the semantic relevance of the learned features, we applied t-SNE~\cite{maaten2008visualizing} to project high-dimensional data to $2$D embedding. We take the output of the last FC layer (see Figure~\ref{fig:architecture}) from the shared network by feeding a limited (but randomly selected) subset of the dataset to extract the embeddings. Further, as the data under consideration is multi-labeled, we identified sets of mutually-exclusive labels (e.g. Indoors vs. Outside) that can be used to color code the data points to visually identify meaningful clusters. Figure~\ref{fig:tsne} provides a visualization for various sets of labels suggesting the network is able to disentangle possible factors of variation that may distinguish a class from the rest in large-scale sensory data. Furthermore, to get better insights in the diversity of the extracted features from each modality, in Figure~\ref{fig:fm}, we visualize the feature maps produced by the first layer of the DPS-Conv layer of modal-specific networks. 

\section{Conclusions}
In this work, we tackled the problem of multi-label behavioral context recognition with deep multi-modal convolutional neural networks. We propose to train an end-to-end model for jointly-learning from low-level sensory data (accelerometer, gyroscope, audio and phone state) of smart devices collected in-the-wild. Our empirical results demonstrated various strategies for feasibly fusing representations learned from different modalities and quantifying their contribution on the predictive performance. We also showed that instance-weighted cross-entropy loss (as also leveraged in~\cite{vaizman2018context}) and regularization schemes enable the model to generalize well on highly imbalanced (sparsely labeled) dataset. Furthermore, we present a slight modification in the proposed network's architecture to handle missing sensors; potentially taking advantage of multi-task learning. We believe, the proposed methodology is generic enough and can be applied to other related problems of learning from multivariate time series. Additionally, potential directions for future work would involve developing techniques to handle imbalanced multi-label data, optimal sensor selection to reduce computation and battery consumption, and incorporating other analogous sensors to further improve the detection rate.

%\section{Acknowledgement}
{\bf \noindent Acknowledgment} \footnotesize{SCOTT (www.scott-project.eu) has received funding from the Electronic Component Systems for European Leadership Joint Undertaking under grant agreement No 737422. This Joint Undertaking receives support from the European Union’s Horizon 2020 research and innovation programme and Austria, Spain, Finland, Ireland, Sweden, Germany, Poland, Portugal, Netherlands, Belgium, Norway.
\\
\noindent Various icons used in the figures are created by Anuar Zhumaev, Tim Madle, Korokoro, Gregor Cresnar, Shmidt Sergey, Hea Poh Lin, Natalia Jacquier, Trevor Dsouza, Adrien Coquet, Alina Oleynik, Llisole, Alena,  AdbA Icons, Jeevan Kumar, Artdabana@Design, lipi, Alex Auda Samora, and Michelle Colonna from the Noun Project.}

\footnotesize{\bibliography{main}}
\bibliographystyle{aaai}
\end{document}